\documentclass[twoside,11pt]{article}

\usepackage{blindtext}

\usepackage{jmlr2e}
\usepackage{amsmath}
\usepackage{bbold}
\usepackage{tikz-cd}

\newcommand{\1}{\mathbb{1}}
\newcommand{\Aut}[1]{\operatorname{Aut}\left(#1\right)}
\newcommand{\ob}[1]{\operatorname{ob}\left(#1\right)}
\newcommand{\mor}[1]{\operatorname{mor}\left(#1\right)}
\newcommand{\dom}[1]{\operatorname{dom}\left(#1\right)}
\newcommand{\cod}[1]{\operatorname{cod}\left(#1\right)}
\renewcommand{\hom}[2]{\operatorname{hom}\left(#1, #2\right)}
\newcommand{\Diff}[1]{\operatorname{Diff}\left(#1\right)}
\newcommand{\nat}[1]{\operatorname{nat}\left(#1\right)}

\usepackage{lastpage}
\jmlrheading{}{2023}{1-\pageref{LastPage}}{12/07}{}{}{Christian Goodbrake}

\ShortHeadings{Unnatural Algorithms in Machine Learning}{Goodbrake}
\firstpageno{1}

\begin{document}

\title{Unnatural Algorithms in Machine Learning}

\author{\name Christian Goodbrake \email christian.goodbrake@oden.utexas.edu \\
       \addr Oden Institute for Computational Engineering and Sciences\\
       University of Texas\\
       Austin, TX 78712, USA
       }

\editor{}

\maketitle

\begin{abstract}%   <- trailing '%' for backward compatibility of .sty file
Natural gradient descent has a remarkable property that in the small learning rate limit, it displays an invariance with respect to network reparameterizations, leading to robust training behavior even for highly covariant network parameterizations. 
We show that optimization algorithms with this property can be viewed as discrete approximations of natural transformations from the functor determining an optimizer's state space from the diffeomorphism group if its configuration manifold, to the functor determining that state space's tangent bundle from this group.
Algorithms with this property enjoy greater efficiency when used to train poorly parameterized networks, as the network evolution they generate is approximately invariant to network reparameterizations.
More specifically, the flow generated by these algorithms in the limit as the learning rate vanishes is invariant under smooth reparameterizations, the respective flows of the parameters being determined by equivariant maps.
By casting this property a natural transformation, we allow for generalizations beyond equivariance with respect to group actions; this framework can account for non-invertible maps such as projections, creating a framework for the direct comparison of training behavior across non-isomorphic network architectures, and the formal examination of limiting behavior as network size increases by considering inverse limits of these projections, should they exist.
We introduce a simple method of introducing this naturality more generally and examine a number of popular machine learning training algorithms, finding that most are unnatural.
\end{abstract}

\begin{keywords}
natural gradient descent, natural transformation, reparameterization invariance, group equivariance
\end{keywords}

\section{Introduction}
\label{sec:intro}

Drawing from the world of information geometry, the ``natural gradient descent'' \citep{amari1998natural, amari2010information} has been proposed and widely studied as an alternative to the traditional gradient descent algorithm for training neural networks.
In this approach, the Fisher information matrix is used to endow the network's parameter space with a Riemannian structure.
With this added structure, the step taken at each update is in the direction of the steepest descent relative to the space of realizable distributions, rather than with respect to the network's parameters. 
The Fisher information matrix is used to ``correct'' the ordinary gradient to account for the the geometry of the network's parameters, and it has been shown that this can improve convergence by accounting for covariance between network parameters analogous to signal whitening \citep{sohl2012natural}.
Remarkably, this structure is invariant under reparameterizations of the network, as the components of the Fisher information matrix transform in a manner complementary to the components of the gradient under reparameterizations, leading to robust training behavior for natural gradient descent even in poorly parameterized networks \citep{pascanu2013revisiting}.
This structure has been studied in much depth, leading to the family of algorithms derived from the information-geometric optimization approach \citep{ollivier2017information}.

Variants of this approach have been developed along numerous lines.
\cite{martens2020new} notes that the natural gradient descent can be viewed as a second order method, placing it alongside other second order optimization methods, for instance those based on Newton's method \citep{karimireddy2018global, nesterov2003introductory}, and the generalized Gauss-Newton matrix \citep{schraudolph2002fast}.
Other methods have been developed to resolve this by approximating the Hessian by only using components of the loss function \citep{kovalev2019stochastic}.
Quasi-Newton methods such as BFGS \citep{international1990bfgs} and its limited memory variant \citep{liu1989limited} seek to limit this computational cost by incrementally updating an estimate of the Hessian, or its inverse at each iteration to avoid having to compute and invert a full Hessian at every iteration.
Methods such as Hessian free \citep{martens2010deep, martens2011learning} and Krylov subspace optimization \citep{mizutani2003iterative, vinyals2012krylov} employ quasi-second order information without having to compute a Hessian matrix, while methods such as Adam \citep{kingma2017adam} and adagrad \citep{duchi2011adaptive} generate diagonal approximations of the second order moment to correct the gradient, essentially diagonally approximating the Fisher information matrix.
Modifications to natural gradient descent such as TONGA \citep{roux2007topmoumoute} introduce low rank approximations to improve the computational cost of computing and inverting the information matrix, achieving greater performance than stochastic gradient descent and minibatch gradient descent both at the level of CPU time and on a per iteration basis.

Given the increasing use of machine learning in scientific computing and other fields outside of data science, we seek to generalize the structure that gives natural gradient descent its reparameterization invariance so that this structure can be extended to settings where information geometry may not be appropriate.
Specifically, we are interested in using machine learning for scientific computing where physical laws provide a Riemannian structure.
We recognize that once these iterative algorithms are applied to any specific neural network, they can be viewed as discrete approximations to a flow on that network's state space, which depends on the algorithm, and that natural gradient descent has the property that the functions determining this flow for each network are the components of a natural transformation.
This general view of naturality provides a clear path for generating new optimization algorithms, for modifying existing methods to provide greater robustness, and a novel framework for network convergence and error analysis.

\section{Prerequisite Mathematics}
\label{sec:math}
As this work is intended to be self contained, all necessary definitions are briefly presented.
These definitions are standard, so we refer the reader to any standard text on category theory such as those by \cite{adkimek2006abstract, mac2013categories, riehl2017category} as well as the seminal paper by \cite{eilenberg1945general} and texts on group theory such as those by \cite{dummit2004abstract, kargapolov1979fundamentals, rotman2012introduction} and the references compiled by \cite{dummitbasic} for further treatment if necessary.
Readers familiar with the basics of category theory, namely categories, automorphisms, functors, natural transformations, and groupoids, along with basic elements of group theory, namely groups, homomorphisms, group actions, and equivariant functions can skip to Section \ref{sec:ml}.

\subsection{Category Theory}
We begin with the abstract notion of a \emph{category}:
\begin{definition}[Category]
A \textbf{category} $\mathsf{C}$ consists of 
\begin{enumerate}
\item A class $\operatorname{ob}\left(\mathsf{C}\right)$ of objects.
\item A class $\operatorname{mor}\left(\mathsf{C}\right)$ of morphisms (sometimes called arrows).
\item A domain or source class function $\operatorname{dom}:\mor{\mathsf{C}}\to\ob{\mathsf{C}}$.
\item A codomain or target class function $\operatorname{cod}:\mor{\mathsf{C}}\to\ob{\mathsf{C}}$.
\item For every three objects $A$, $B$ and $C$, a binary operation $\hom{A}{B} \times \hom{B}{C} \to \hom{A}{C}$ called composition of morphisms. Here $\hom{A}{B}$ denotes the subclass of morphisms $f$ in $\mor{\mathsf{C}}$ such that $\dom{f} = A$ and $\cod{f} = B$. Morphisms in this subclass are written $f: A \to B$, and the composite of $f : A \to B$ and $g : B \to C$ is often written as $g \circ f$ or $gf$.
\end{enumerate}
Further, the following conditions must be satisfied:
\begin{enumerate}
\item Associativity: if $f : A \to B$, $g : B \to C$ and $h : C \to D$ then $h \circ \left(g \circ f\right) = \left(h \circ g\right) \circ f$.
\item Identities: for every object $A$, there exists a morphism $\1_A : A \to A$ (some authors write $\operatorname{id}_A$) called the identity morphism for $A$, such that every morphism $f : B \to A$ satisfies $\1_A \circ f = f$, and every morphism $g : A \to B$ satisfies $g \circ \1_A = g$.
\end{enumerate}
\end{definition}

Much of the richness of category theory comes from the explicit consideration of the morphisms, with particular special classes of morphisms standing out, such as epimorphisms, and monomorphisms. 
Here we introduce the few needed for this construction:

\begin{definition}[Isomorphism]
A morphism $f:A\to B$ is an \textbf{isomorphism} iff there exists a morphism $f^{-1}:B\to A$ such that $f\circ f^{-1} = \1_B$ and $f^{-1} \circ f = \1_A$. Objects $A$ and $B$ are then said to be \textbf{isomorphic}.
\end{definition}

Note that the question of the existence of isomorphism depends not only on the objects $A$ and $B$, but also the category containing $f$. 
Therefore, when speaking of two isomorphic objects, it is important to note the category in which the isomorphism exists if it is not clear by context, since two objects may be isomorphic in one category, but not in another e.g. ``$A$ and $B$ are isomorphic as groups, but not as rings.''

\begin{definition}[Endomorphism]
A morphism $f$ is an \textbf{endomorphism} iff $\operatorname{dom}\left(f\right) = \operatorname{cod}\left(f\right)$.
\end{definition}

Combining the previous two ideas, we obtain the notion of an automorphism.

\begin{definition}[Automorphism]
An endomorphism that is also an isomorphism is an \textbf{automorphism}.
\end{definition}

Next, we consider ``maps between categories,'' called \emph{functors}.
\begin{definition}[Functor]
Let $\mathsf{C}$ and $\mathsf{D}$ be categories. A (covariant) \textbf{functor} $F$ is a mapping that
\begin{enumerate}
\item associates each object  $A$ in $\mathsf{C}$ to an object $F\left(A\right)$ in $\mathsf{D}$
\item associates each morphism $f: A \to B$ in $\mathsf{C}$ to a morphism $F\left(f\right): F\left(A\right) \to F\left(B\right)$ in $\mathsf{D}$ such that the following two conditions hold:
\begin{enumerate}
\item $F\left(\1_A\right) = \1_{F\left(A\right)}$ for every object $A$ in $\mathsf{C}$.
\item $F\left(g\circ f\right) = F\left(g\right)\circ F\left(f\right)$ for all morphisms $f:A\to B$ and $g: B\to C$ in $\mathsf{C}$.
\end{enumerate}
\end{enumerate}

\end{definition}

There also exist \emph{contravariant} functors that nearly satisfy the above axioms except they reverse the order of morphism composition, i.e. for a contravariant functor $F$, $F\left(g\circ f\right) = F\left(f\right)\circ F\left(g\right)$ for all morphisms $f:A\to B$ and $g: B\to C$ in $\mathsf{C}$, where $F\left(f\right) : F\left(B\right)\to F\left(A\right)$, and $F\left(g\right):F\left(C\right)\to F\left(B\right)$.

To relate functors to one another, we define an \emph{infranatural transformation}

\begin{definition}[Infranatural Transformation]
Let $F,H:\mathsf{C}\to \mathsf{D}$ be two functors. 
An \textbf{infranatural transformation} $\eta$ from $F$ to $H$ is a family of morphisms with the following structure:
\begin{itemize}
\item For every $A\in \ob{\mathsf{C}}$, the infranatural transformation associates a morphism $\eta_A:F\left(A\right)\to H\left(A\right)$ between objects in $\mathsf{D}$. The morphism $\eta_A$ is known as the component of $\eta$ at $A$.
\end{itemize}
\end{definition}

Infranatural transformations do not necessarily respect the categorical structure of the functors they relate; imposing this compatibility yields the definition of a \emph{natural transformation} between functors, which can be thought of as a ``morphism of functors,'' in that it transforms one functor into another while respecting the underlying categorical structures involved.
In fact, one can define a category whose objects are functors and whose morphisms are precisely natural transformations, so the conception of a natural transformation as a morphism of functors can be taken quite literally.

\begin{definition}[Natural Transformation]
Let $F,H:\mathsf{C}\to \mathsf{D}$ be two functors. 
A \textbf{natural transformation} $\eta$ from $F$ to $H$ is an infranatural transformation from $F$ to $H$ satisfying the following commutativity requirement:
\begin{itemize}
\item Components must be such that for every morphism $f:A\to B$ in $\mathsf{C}$ we have $\eta_B \circ F\left(f\right) = H\left(f\right) \circ \eta_A$, i.e. the following diagram must commute:
\end{itemize}
\begin{center}
\begin{tikzcd}
A \arrow{d}{f} &F\left(A\right) \arrow{r}{\eta_A} \arrow[swap]{d}{F\left(f\right)} & H\left(A\right) \arrow{d}{H\left(f\right)} \\%
B &F\left(B\right) \arrow{r}{\eta_B}& H\left(B\right)
\end{tikzcd}
\end{center}
\end{definition}

Historically speaking, natural transformations have been the primary motivation behind the development of category theory, with the definitions of categories and functors being necessary steps toward properly defining natural transformations.

Given an infranatural transformation $\eta$, one might ask if we can make it natural by restricting the source category $\mathsf{C}$ to some subcategory. 
This is possible, and the largest such subcategory is called the \emph{naturalizer} of $\eta$, denoted $\nat{\eta}$.
Note briefly that the naturalizer of an infranatural transformation can be obtained by excluding only morphisms from consideration, as if $f:A\to A$ is the identity morphism for the object $A\in\ob{\mathsf{C}}$, then the commutativity requirement reduces to the trivial condition
\begin{equation*}
\eta_A \circ \1_{F\left(A\right)} = \1_{H\left(A\right)} \circ \eta_A.
\end{equation*} 
Therefore, every object in the source category can be included in the naturalizer, along with at least its identity morphism.

\subsection{Group Theory}
Next, we take a slight detour into the language of group theory.
Groups are the natural language of symmetry, and are typically the first type of object one considers in an abstract algebra course.
\begin{definition}[Group]
A \textbf{group} $\left(G,*\right)$ is a set $G$ equipped with a binary operator $*:G\times G\to G$ satisfying
\begin{enumerate}
\item Associativity: $\left(a*b\right)*c = a*\left(b*c\right),\, \forall a,b,c,\in G$.
\item Identity: $\exists$ $e\in G$ such that $e*g=g*e=g$ $\forall g\in G$.
\item Inverse: $\forall g\in G,\, \exists g^{-1}\in G$ such that $g*g^{-1}=g^{-1}*g=e$.
\end{enumerate}
\end{definition}
When the operator $*$ is commutative, it is often called ``addition,'' and when it is not, it is often called ``multiplication.'' 
When a group's multiplication is obvious by context (as it usually is), the group will simply be denoted as $G$, i.e. the multiplication symbol will be omitted.

With this, we can see that we can endow the set of automorphisms of an object with the structure of a group by defining the group multiplication to be morphism composition. 
The group axioms follow trivially from the associativity of morphism composition, the existence of identity morphisms, and the invertibility of automorphisms.
We denote the automorphism group of an object $A$ as $\Aut{A}$.

While this definition fully characterizes groups from an algebraic standpoint, much richness can be gained by considering the relationships between groups and other groups.
First, we consider ``structure preserving maps'' between groups.

\begin{definition}[Group Homomorphism]
A map $\varphi:G\to H$ between two groups $\left(G,*\right)$ and $\left(H,\cdot\right)$ is a \textbf{homomorphism} if it satisfies
\begin{enumerate}
\item $\varphi\left(g_2*g_1\right)=\varphi\left(g_2\right)\cdot\varphi\left(g_1\right),\forall g_1,g_2 \in G$.
\end{enumerate}
\end{definition}
In short, a group homomorphism preserves multiplicative structure of a group.
With this, it is clear to see that we can define a category of groups, whose objects are groups and whose morphisms are group homomorphisms.
It is also useful to consider the relationship between groups and other objects:
\begin{definition}[Group Action]
A group $G$ acts on an object $M$ (on the left) through a (left) \textbf{group action}, which to each $g\in G$ associates a map $\mu_g : M\to M$  satisfying the following axioms:
\begin{enumerate}
\item Identity: $\mu_e =\1_M$.
\item Compatibility: $\mu_{g_2}\circ\mu_{g_1} = \mu_{g_2*g_1},\forall g_1,g_2\in G$.
\end{enumerate}
A right group action is similar, but with the order of actions in the compatibility reversed.
\end{definition}

Because $\left(g_2*g_1\right)^{-1}=g_1^{-1}*g_2^{-1}$, right actions can be converted into left actions by composing with the inverse operation of the group. A right action can also be considered as the left action of the opposite group $G^{op}$.
An object $M$ equipped with an action by a group $G$ is called a $G$-object.
Relating $G$-objects, we have

\begin{definition}[$G$-equivariance]
A map $f:M\to N$ between $G$-objects $M$ and $N$ with respective $G$-actions $\mu$ and $\nu$ is \textbf{$G$-equivariant} if
\begin{equation*}
f\circ \mu_g = \nu_g\circ f, \, \forall g\in G.
\end{equation*}
\end{definition}
Essentially $G$-equivariant functions ``commute'' with the $G$-actions on the domain and codomain in the sense that for all $g\in G$, the following diagram commutes:
\begin{center}
\begin{tikzcd}
M \arrow{r}{f} \arrow[swap]{d}{\mu_g} & N \arrow{d}{\nu_g} \\%
M \arrow{r}{f}& N
\end{tikzcd}
\end{center}

Finally, we note that there is another connection between categories and groups.
Namely, we first define the notion of a \emph{groupoid}.
\begin{definition}[Groupoid]
A \textbf{groupoid} is a category in which every morphism is an isomorphism.
\end{definition}
We see that a group can be considered formally as a groupoid with one object, say $\bullet$, where the morphisms $\hom{\bullet}{\bullet}$ are defined to be the group elements, and morphism composition is defined to be the group multiplication. 
The identity morphism is simply the group's identity element.
When we consider a group $G$ in this manner, we will refer to its associated one object category as $\mathsf{BG}$.
\subsection{Alternative Characterizations of Group Actions}
With the machinery developed thus far, we can cast group actions as two different but equivalent types of objects.
Beginning with the group theoretic approach:
\begin{definition}[Group Action as Homomorphism]
A left group action of a group $G$ on an object $M$ is given by a homomorphism $\varphi:G\to \operatorname{Aut}\left(M\right)$, with the action itself being
\begin{equation*}
\mu_g = \varphi\left(g\right), \forall g\in G.
\end{equation*}
\end{definition}

We see that the definitions of the automorphism group and homomorphisms exactly encode the group action axioms.
Alternatively, we can consider a group as a category (i.e. a one object groupoid).
In this case, we can characterize a group action as a functor:
\begin{definition}[Group Action as Functor]
A left group action of a group $G$ on an object $M$ in a category $\mathsf{C}$ is given by a functor $K:\mathsf{BG}\to \mathsf{C}$, with $K\left(\bullet\right)=M$ and
\begin{equation*}
\mu_g = K\left(g\right), \forall g\in G.
\end{equation*}
\end{definition}

These two presentations are equivalent, though we will see that the latter is more readily generalizable.

\subsection{Group Equivariance and Naturality}
The key component driving this work is the fact that we are not concerned with equivariant maps between objects with arbitrary group actions, but rather that the group actions we will examine are determined from a single common group action by a set of functors.
This gives us a reasonable notion of symmetric maps between domains and codomains that are determined by functors acting on a common object possessing its own group action.
When this common object possesses some symmetry, we want our map to respect the effects of this symmetry on its domain and codomain.

To begin, we consider a map $\eta_A:F\left(A\right)\to H\left(A\right)$ in category $\mathsf{D}$, where $A$ is a $G$-object in a category $\mathsf{C}$.
We then consider the category $\mathsf{BG}$ associated with $G$.
The action on $A$ is determined by a functor $K$ from $\mathsf{BG}$ to $\mathsf{C}$, and this functor can be composed with $F$ and $H$.
These composite functors, $F\circ K$ and $H\circ K$ then determine actions on $F\left(A\right)$ and $H\left(A\right)$.

Next, consider an infranatural transformation from $F\circ K$ to $H\circ K$. 
Because $\mathsf{BG}$ has only one object, this transformation has only one component of the form $\eta_A:F\left(A\right)\to H\left(A\right)$, which we take to be our map.
Demanding that this map be the sole component of a natural transformation yields the condition
\begin{equation*}
\eta_A\circ \left(F\circ K\right)\left(g\right)=\left(H\circ K\right)\left(g\right) \circ \eta_A,\, \forall g\in \left(\mor{\mathsf{BG}}=G\right).
\end{equation*}

This is precisely the statement that $\eta_A$ is a $G$-equivariant map, so we see that any map that is equivariant with respect to the functorially induced group actions can be considered the sole component of a natural transformation. 
All of the above taken together is essentially Example 1.3.9 in the text by \cite{riehl2017category}.

We note that even if an explicit $G$-action on $A$ is not given, there is an obvious $\Aut{A}$-action on $A$ that can always be used. In fact, thinking of group actions as homomorphisms into automorphism groups, the kernel of any $G$-action acts trivially (since it maps to the identity), and the image of any $G$-action is a subgroup of $\Aut{A}$, so considering the $\Aut{A}$-action on $A$ is in some sense considering the ``biggest'' symmetry possible.
This of course requires defining a category for $A$; different symmetry conditions on $f$ can be obtained depending on what automorphisms are considered in $\Aut{A}$.

With this, we can expand our categorical perspective, and rather than considering the single object $A$ as an $\Aut{A}$-object, we consider the entire category containing $A$.
Given this category, we can define its \emph{automorphism groupoid}
\begin{definition}[Automorphism Groupoid of a Category]
Let $\mathsf{C}$ be a category. The \textbf{automorphism groupoid} $\Aut{\mathsf{C}}$ of $\mathsf{C}$ is the subcategory of $\mathsf{C}$ whose morphisms consist only of the automorphisms in $\mathsf{C}$. 
\end{definition}
Note there is an obvious inclusion functor mapping $\Aut{\mathsf{C}}\to \mathsf{C}$, since the objects of these two categories are the same, and the morphisms of $\Aut{\mathsf{C}}$ are also morphisms in $\mathsf{C}$.
This inclusion functor simultaneously captures the actions of each group $\Aut{A}$ on each object $A$ in $\mathsf{C}$.

With this, we recognize that if we can associate a function for every object in $\mathsf{C}$ that satisfies our symmetry requirement with respect to the action of its own automorphism group, then these functions are the components of a natural transformation between functors $F,H:\Aut{\mathsf{C}}\to \mathsf{D}$.

\section{Machine Learning Training Algorithms}
\label{sec:ml}
Before we can apply the preceding framework to the analysis of machine learning training algorithms, we first must consider these algorithms in an appropriate limiting sense.
We consider iterative optimization algorithms depending a learning rate parameter $\xi$ training a network with parameters $\theta^i$, with $i\in\{1,...,N\}$.
Any given choice of parameters $\theta^i$ defines a function, and the collection of all such functions forms a smooth manifold $\mathcal{M}$ called the ``configuration manifold'' of the network, which locally is diffeomorphic to $\mathbb{R}^N$.

Two networks $f_{NN_1}\left(x,\theta^i\right)$ and $f_{NN_2}\left(x,\bar{\theta}^i\right)$ depending on the inputs $x$ and the respective sets of parameters $\theta^i$ and $\bar{\theta}^i$ are then isomorphic as neural networks if for each set of parameters $\theta^i$, there is a corresponding set of parameters $\bar{\theta}^i$ such that $f_{NN_1}\left(x,\theta^i\right)=f_{NN_2}\left(x,\bar{\theta}^i\right)$ for all inputs $x$, and likewise there exists $\theta^i$ for every choice of $\bar{\theta}^i$ such that these functions are equal.
Iteratively training a neural network amounts to navigating through the configuration manifold until we find a set of parameters yielding a function that is optimal for our purposes.
We define a category of neural networks whose objects are the configuration manifolds of neural networks, and whose morphisms are diffeomorphisms.
As such, this category is a groupoid, since diffeomorphisms are invertible.
We then take $\mathsf{NNet}$ to be the skeleton of the previously described category, which amounts to restricting its objects to a single object for each isomorphism class, and identifying a single fixed isomorphism from that object to each other object in its isomorphism class with its identity morphism.
This simply identifies isomorphic networks, and turns every isomorphism into an automorphism.
We do this so that the category we work with is its own automorphism groupoid; it is also clearly a subcategory of $\mathsf{Man}$, the category of smooth manifolds, hence any functor from $\mathsf{Man}$ can be precomposed by the inclusion functor $\mathsf{NNet}\to\mathsf{Man}$ taking this neural network category into the category of smooth manifolds to yield a composite functor with $\mathsf{NNet}$ as its source.

For this analysis, we consider a training algorithm as two independent components.
First, let the size of the parameter update converge to zero in an appropriate limit as the learning rate $\xi\to 0$, (recognizing that hyperparameters may have to be changed as functions of $\xi$ to ensure convergence).
In this case, the action of this algorithm converges to an ordinary differential equation for the parameters $\theta^i$ generating a flow on $\mathcal{M}$, for example gradient flow, or in more sophisticated treatments of natural gradient descent, the more general IGO flow \citep{ollivier2017information}.
Depending on the optimizer, this ODE will be of various orders and forms; to obtain a unified form, we write this equation as a first order ODE for an optimizer state.
To do this, suppose that the ODE in question can be written explicitly, i.e. it can locally be written in the form
\begin{equation*}
\frac{d^n\theta^i}{d\xi^n} = \Psi\left(\frac{d^{n-1}\theta^i}{d\xi^{n-1}},\frac{d^{n-2}\theta^i}{d\xi^{n-2}},...,\frac{d\theta^i}{d\xi},\theta^i\right).
\end{equation*}
Defining the state as the following tuple
\begin{equation*}
s = \{\frac{d^{n-1}\theta^i}{d\xi^{n-1}},\frac{d^{n-2}\theta^i}{d\xi^{n-2}},...,\frac{d\theta^i}{d\xi},\theta^i\},
\end{equation*}
the ODE can be rewritten as a first order system
\begin{equation*}
\dot{s} = \eta_{\mathcal{M}}\left(s\right).
\end{equation*}
The state space itself, i.e. $\mathcal{S_M}$ where $s\in\mathcal{S_M}$ is the value of an endofunctor $\mathsf{Man}\to\mathsf{Man}$ applied to $\mathcal{M}$ \citep{kolar2013natural}.
This is evidenced by the fact that if the parameters of one network were written as functions of the parameters of another network, i.e.
\begin{equation*}
\bar{\theta}^i = g^i\left(\theta^j\right),
\end{equation*}
then this induces a map between states through the chain rule, i.e.
\begin{equation*}
\frac{d^k\bar{\theta}^i}{d\xi^k} = \frac{d^k}{d\xi^k}\left(g^i\left(\theta^j\right)\right)
\end{equation*}
can be written entirely in terms of the partial derivatives of $g^i$ and the derivatives of $\theta^i$, the latter of which comprise the state.
In short, a map between configuration manifolds induces a map between the corresponding state spaces, therefore the state space is the value of a functor. 

To recover the optimization algorithm from its limiting ODE, one simply has to apply a numerical integration scheme to the ODE. 
This approach offers much flexibility, as a training ODE with desirable properties can be chosen, and then an integration scheme can be chosen to take advantage of particular additional structures that may be present, generating a specialized training algorithm.

It is important to note that $T\mathcal{S}$, the space containing $\dot{s}$, is determined from the state space $\mathcal{S}$ by the tangent bundle functor $T:\mathsf{Man}\to\mathsf{Man}$, so by composition we have functors determining the spaces for $s$ and $\dot{s} = \eta_{\mathcal{M}}\left(s\right)$, both of which we precompose by the inclusion functor mapping our network automorphism groupoid $\mathsf{NNet}$ into the category of smooth manifolds $\mathsf{Man}$.
Because our training algorithm associates such a function $\eta_{\mathcal{M}}$ to every network, we recognize that this small learning rate limit converts a training algorithm into an infranatural transformation from the functor yielding $\mathcal{S}$ to the functor yielding $T\mathcal{S}$.
The reparameterization invariance enjoyed by natural gradient descent is tantamount to the stronger statement that this transformation is actually a natural transformation.

\begin{definition}[Natural Training]
For every neural network with configuration manifold $\mathcal{M}$, let a training algorithm converge to an ODE of the form $\dot{s} = \eta_{\mathcal{M}}\left(s\right)$ as the learning rate approaches $0$. The training algorithm is \textbf{natural} if each function $\eta_{\mathcal{M}}$ is $\Diff{\mathcal{M}}$-equivariant with respect to the functorially induced actions on $\mathcal{S}$ and $T\mathcal{S}$, i.e. the functions $\eta_{\mathcal{M}}$ are the components of a natural transformation between the functors $F,H:\mathsf{NNet}\to\mathsf{Man}$ determining $\mathcal{S}$ and $T\mathcal{S}$ from $\Aut{\mathcal{M}}=\Diff{\mathcal{M}}$.
\end{definition}
In this sense, we can identify natural machine learning training algorithms as discrete approximations of a natural transformation between functors, shown in the following commutative diagram:

\begin{center}
\begin{tikzcd}
\mathcal{M} \arrow{d}{g} &\mathcal{S_M} \arrow{r}{\eta_{\mathcal{M}}} \arrow[swap]{d}{\mathcal{S_M}\left(g\right)} & T\mathcal{S_M} \arrow{d}{T\mathcal{S_M}\left(g\right)} \\%
\mathcal{M} &\mathcal{S_M} \arrow{r}{\eta_{\mathcal{M}}}& T\mathcal{S_M}
\end{tikzcd}
$, \quad \forall g\in \Diff{\mathcal{M}},$
\end{center}
which is a general natural transformation with $F\left(\mathcal{M}\right) = \mathcal{S_M}$, $H\left(\mathcal{M}\right) = T\mathcal{S_M}$, $F\left(g\right) = \mathcal{S_M}\left(g\right)$, and $H\left(g\right) = T\mathcal{S_M}\left(g\right)$, and all morphisms, $g$, are automorphisms.

Equivalently, if we consider the functions $\eta_{\mathcal{M}}$ determining the flows $\dot{s}=\eta_{\mathcal{M}}\left(s\right)$ as the components of an infranatural transformation $\eta:\mathsf{Man}\to \mathsf{Man}$, we call a training algorithm ``natural'' if its naturalizer is $\mathsf{NNet}$, i.e. $\nat{\eta}=\mathsf{NNet}$.

\section{Examples}
Most of the most common training algorithms used in machine learning are not natural in the sense previously defined.
\subsection{Counterexamples}
\begin{example}[Gradient Descent]
The simplest training algorithm we consider is gradient descent, in which parameters $\theta^i$ are updated according to 
\begin{equation*}
\Delta \theta^i = - \xi \frac{\partial L}{\partial \theta^i},
\end{equation*}
where $\xi$ is a step size hyperparameter.
Dividing through by $\xi$ and taking the limit as $\xi\to 0$, we see that gradient descent converges to \emph{gradient flow}.
\begin{equation}\label{eqn:grad_flow}
\frac{d\theta^i}{d\xi} = -\frac{\partial L}{\partial \theta^i}.
\end{equation}

If we introduce a diffeomorphism $\theta^i\to\bar{\theta}^i$, and transform equation \ref{eqn:grad_flow} into the barred parameters, we obtain
\begin{equation}\label{eqn:transformedgradflow}
\frac{d\bar{\theta}^i}{d\xi} = -\sum_{j,k}^N\frac{\partial \bar{\theta}^i}{\partial \theta^j}\frac{\partial \bar{\theta}^k}{\partial \theta^j}\frac{\partial L}{\partial \bar{\theta}^k},
\end{equation}
whereas, had we applied the diffeomorphism first and then computed the evolution, we would simply have obtained
\begin{equation}\label{eqn:untransformedgradflow}
\frac{d\bar{\theta}^i}{d\xi} = -\frac{\partial L}{\partial \bar{\theta}^i}.
\end{equation}
As such, we see that gradient descent is not natural, since it converges to gradient flow, which is not a component of a natural transformation, since it does not obey the commutativity requirement of natural transformations.
This unnaturality can be seen by the fact that the left hand side of equation \ref{eqn:grad_flow} is a tangent vector field, i.e. a section of the tangent bundle $T\mathcal{M}$, while the right hand side is a section of the cotangent bundle $T^*\mathcal{M}$. To equate these two objects, one must employ an isomorphism dependent on the basis induced by the network's parameterization, which despite being an isomorphism, is unnatural.
Note that this example not only shows that gradient descent is unnatural, but so is any other algorithm that converges to gradient flow as the learning rate approaches zero.

We can compute the naturalizer of this transformation, i.e. the set of reparameterizations that render equations \ref{eqn:transformedgradflow} and \ref{eqn:untransformedgradflow} equal. 
Specifically, denoting $Q^i_j=\frac{\partial \bar{\theta}^i}{\partial \theta^j}$, we require
\begin{equation}\label{eqn:grad_flow_nat}
\sum_j^N Q^i_j Q^k_j = \delta^{ik}
\end{equation}
where $\delta^{ik}$ is the Kronnecker delta.
This implies that $Q^i_j$ is an orthogonal matrix with respect to the standard inner product in $\mathbb{R}^n$.
As this is a reparameterization, and hence is bijective, this implies by the Mazur–Ulam theorem that the reparameterization is affine, i.e.
\begin{equation}\label{eqn:reparameterization}
\bar{\theta}^i = \sum_j^N Q^i_j\theta^j + c^i
\end{equation}
for constant vector $c^i\in \mathbb{R}^N$ and constant orthogonal matrix $Q^i_j$.
We therefore recognize that the naturalizer for gradient descent is the category $\mathsf{NNet}$ with the morphisms restricted to Euclidean motions.
A general configuration manifold with coordinates $\theta^i$ will not necessarily be globally compatible with this Euclidean structure, in which case reparameterizations of the form of equation \ref{eqn:reparameterization} will not necessarily exist for all choices of $Q^i_j$ and $c^i$. 
However, all morphisms in $\nat{\eta}$ will have this form, and every object in the naturalizer will at least have the special case $c^i=0$ and $Q^i_j=\delta^i_j$, this being the identity morphism for each object.
\end{example}

\begin{example}[Nesterov's Accelerated Gradient]
As shown by \cite{su2015differential}, in the small learning rate limit, Nesterov's Accelerated Gradient converges to 
\begin{equation}\label{eqn:nag}
\frac{d^2\theta^i}{d\xi^2} = -\frac{3}{\xi}\frac{d\theta^i}{d\xi} - \frac{\partial L}{\partial \theta^i}.
\end{equation}
Similarly to gradient flow, this equation employs a hidden unnatural isomorphism to be well defined, which is revealed by applying a diffeomorphism $\theta^i\to \bar{\theta}^i$:
\begin{equation*}
\sum_j^N\frac{\partial \theta^i}{\partial \bar{\theta}^j}\frac{d^2\bar{\theta}^j}{d\xi^2} = -\frac{3}{\xi}\sum_j^N\frac{\partial \theta^i}{\partial \bar{\theta}^j}\frac{d\bar{\theta}^j}{d\xi} - \sum_j^N\frac{\partial \bar{\theta}^j}{\partial \theta^i} \frac{\partial L}{\partial \bar{\theta}^j}.
\end{equation*}
Inverting $\frac{\partial \theta^i}{\partial \bar{\theta}^j}$ yields the new evolution equation 
\begin{equation*}
\frac{d^2\bar{\theta}^i}{d\xi^2} = -\frac{3}{\xi}\frac{d\bar{\theta}^i}{d\xi} - \sum_{j,k}^N\frac{\partial \bar{\theta}^i}{\partial \theta^k}\frac{\partial \bar{\theta}^j}{\partial \theta^k} \frac{\partial L}{\partial \bar{\theta}^j}.
\end{equation*}
which is not equal to equation \ref{eqn:nag} with $\theta^i$ replaced with $\bar{\theta}^i$, and hence this training flow is not a natural transformation, and Nesterov's accelerated gradient is not natural.
We also see the same naturality condition, equation \ref{eqn:grad_flow_nat}, that we saw for gradient descent, and as such, the naturalizer for this algorithm is the same for gradient descent, i.e. Euclidean transformations of the form
\begin{equation*}
\bar{\theta}^i = \sum_j^N Q^i_j\theta^j + c^i
\end{equation*}
for constant vector $c^i\in \mathbb{R}^N$ and constant orthogonal matrix $Q^i_j$.
\end{example}

\begin{example}[Adam]
Adam enjoys invariance under diagonal rescaling of gradients in a certain sense, which can be seen by observing how Adam acts on one parameter, say $\theta$.
As all parameters are treated independently, this is sufficient as the construction immediately generalizes to the multivariate case.
Adam estimates of the first and second moments of the loss function by keeping exponential moving averages.
\begin{equation*}
m =  \left(1-\beta_1\right)\sum_{i=1}^t\beta_1^{t-i} \frac{\partial L}{\partial \theta},\quad v = \left(1-\beta_2\right)\sum_{i=1}^t\beta_2^{t-i} \frac{\partial L}{\partial \theta}^2
\end{equation*}
When these moments are stationary, the expectations of the estimates are equal to scaled versions of the expectations of the components of the gradient, and their squares i.e. for each $\theta$ we have
\begin{equation*}
\mathbb{E}\left(m\right) =\left(1-\beta_1^t\right) \mathbb{E}\left(\frac{\partial L}{\partial \theta}\right), \quad \mathbb{E}\left(v\right) =\left(1-\beta_2^t\right) \mathbb{E}\left(\left(\frac{\partial L}{\partial \theta}\right)^2\right),
\end{equation*}
hence the introduction of the bias corrected $\hat{m}$ and $\hat{v}$.
\begin{equation*}
\hat{m} = \frac{m}{\left(1-\beta_1^t\right)}, \quad \hat{v} =\frac{v}{\left(1-\beta_2^t\right)}
\end{equation*}
Taking the limit of the $\theta$ update equation as the learning rate goes to zero yields an equation of the form
\begin{equation*}
\frac{d\theta}{d\xi} = -\frac{\hat{m}}{\sqrt{\hat{v}}}.
\end{equation*}
Because of the square root appearing in the denominator, if we rescale the gradient $\frac{\partial L}{\partial \theta}\to \gamma \frac{\partial L}{\partial \theta}$, with $\gamma>0$, then $\hat{m}\to \gamma\hat{m}$ and $\hat{v}\to \gamma^2\hat{v}$, hence $\frac{d\theta}{d\xi}$ is unchanged.
For $\gamma<0$, a sign change is introduced.
However, if the rescaling of $\frac{\partial L}{\partial \theta}$ is caused to the reparameterization $\theta \to \gamma^{-1}\theta$, while the right hand side is unchanged, the left hand side transforms as 
\begin{equation*}
\frac{d\theta}{d\xi}\to\gamma^{-1}\frac{d\theta}{d\xi},
\end{equation*}
violating naturality even over this restricted set of reparameterizations. 
Were it not for the square root, this update would transform properly, and Adam would enjoy naturality with respect to one dimensional reparameterizations applied to its parameters individually.
Absent this correction, this one dimensional behavior is only preserved for $|\gamma|=1$, i.e. reparameterizations that preserve the norm $d\theta^2$.
Further considering the multivariate case, because the gradients are preconditioned by an essentially arbitrary diagonal matrix depending on the loss function (whose diagonal entries are the values $\sqrt{\hat{v}^i}$), we do not retain equivariant behavior under general norm-preserving i.e. orthogonal transformations; the only reparameterizations that generate equivariant behavior are ones where $|\frac{\partial L}{\partial \theta^i}| = |\frac{\partial L}{\partial \bar{\theta}^i}|$, for each parameter $\theta^i$ while preserving the arbitrary diagonal structure.
As such, the reparameterization must be of the form
\begin{equation*}
\bar{\theta}^i =  \sum_j^N P^i_j \theta^j + c^i,
\end{equation*}
for signed permutation matrix $P^i_j$ and constant vector $c^i$.
\end{example}

\begin{example}[Newton's Method]
The naturality of Newton's Method depends on whether or not the Hessian is computed using an affine connection.

The Newton update step is as follows:
\begin{equation*}
\Delta \theta^i =- \xi \sum_j^N \left(H^{-1}\right)^{ij}\frac{\partial L}{\partial \theta^j},
\end{equation*}
where $H_{ij}=\frac{\partial^2L}{\partial \theta^i\partial\theta^j}$ is the Hessian.
Clearly, the corresponding limiting flow is
\begin{equation*}
\frac{d \theta^i}{d\xi} =- \sum_j^N \left(H^{-1}\right)^{ij}\frac{\partial L}{\partial \theta^j}.
\end{equation*}
Performing a reparameterization $\theta^i\to \bar{\theta}^i$ transforms the Hessian as follows
\begin{equation}\label{eqn:Hessian}
\frac{\partial^2L}{\partial\theta^i\partial\theta^j} \to\frac{\partial^2L}{\partial\bar{\theta}^i\partial\bar{\theta}^j}= \sum_l^N \frac{\partial^2\theta^l}{\partial \bar{\theta}^i \partial \bar{\theta}^j}\frac{\partial L}{\partial \theta^l}+\sum_{l,p}^N \frac{\partial\theta^p}{\partial \bar{\theta}^i}\frac{\partial ^2L}{\partial\theta^p\partial\theta^l}\frac{\partial\theta^l}{\partial \bar{\theta}^j}.
\end{equation}
However, for naturality, we need 
\begin{equation*}
H_{ij}\to \bar{H}_{ij} =\sum_{k,l}^N \frac{\partial \theta^k}{\partial \bar{\theta}^i}H_{kl}\frac{\partial \theta^l}{\partial \bar{\theta}^j}.
\end{equation*}
Observe that merely taking a matrix of partial derivatives is unnatural due to the first term in equation \ref{eqn:Hessian}.
This problem is precisely fixed by the covariant derivative, which uses the affine connection to yield a properly transforming Hessian.
Therefore, if we have an affine connection, and compute the Hessian using the covariant derivative, we obtain a natural transformation that can be discretized into a training algorithm with a suitable integration scheme:
\begin{align*}
\frac{d \theta^i}{d\xi} =- \sum_j^N \left(H^{-1}\right)^{ij}\frac{\partial L}{\partial \theta^j}\to \sum_j^N \frac{\partial \theta^i}{\partial \bar{\theta}^j}\frac{d \bar{\theta}^j}{d\xi} =-  \sum_{j,l}^N \left(H^{-1}\right)^{ij}\frac{\partial\bar{\theta}^l}{\partial\theta^j}\frac{\partial L}{\partial \bar{\theta}^l}\\
\Rightarrow\frac{d \bar{\theta}^i}{d\xi} =-  \sum_j^N\left(\bar{H}^{-1}\right)^{ij}\frac{\partial L}{\partial \bar{\theta}^j}.
\end{align*}

If we compute the naturalizer of Newton's method using standard partial derivatives, we see that the first term in equation \ref{eqn:Hessian} must vanish, i.e. we require 
\begin{equation*}
\sum_l^N \frac{\partial^2\theta^l}{\partial \bar{\theta}^i \partial \bar{\theta}^j}\frac{\partial L}{\partial \theta^l}=0.
\end{equation*}
Since we desire this for all loss functions $L$, we require 
\begin{equation*}
 \frac{\partial^2\theta^l}{\partial \bar{\theta}^i \partial \bar{\theta}^j}=0,
\end{equation*}
i.e. the reparameterization must be affine:
\begin{equation*}
\theta^i = A^i_j\bar{\theta}^j + c^i,
\end{equation*}
for constant matrix $A^i_j$ and vector $c^i$.
\end{example}

The following table summarizes the equivariance groups of the previous algorithms:
\begin{center}
\begin{tabular}{ | c | c | c | }
\hline
Algorithm & Reparameterization & Equivariance Group\\
\hline
 Gradient Descent & $\bar{\theta}^i = \sum_j^N Q^i_j\theta^j + c^i$   & $E\left(N\right)=O\left(N\right) \ltimes T\left(N\right)$  \\ 
 Nesterov Accelerated Gradient & $\bar{\theta}^i = \sum_j^N Q^i_j\theta^j + c^i$ & $E\left(N\right)=O\left(N\right) \ltimes T\left(N\right)$ \\  
 Adam & $\bar{\theta}^i = \sum_j^N P^i_j\theta^j + c^i$ & $B_N \ltimes T(N)$ \\
 Newton's method & $\bar{\theta}^i = \sum_j^NA^i_j\theta^j + c^i$ & $\text{Aff}\left(N,\mathbb{R}\right)=GL\left(N,\mathbb{R}\right) \ltimes T\left(N\right)$\\
 \hline
 \end{tabular}
\end{center}
where above, $E\left(N\right)$ is the Euclidean group, $O\left(N\right)$ is the orthogonal group, $T\left(N\right)$ is the translation group, $B_N$ is the hyperoctahedral group (which is the symmetry group of the $N$ dimensional hypercube), $\text{Aff}\left(N,\mathbb{R}\right)$ is the real affine group, and $GL\left(N,\mathbb{R}\right)$ is the real general linear group.
For the representations of the natural reparameterizations appearing in this table, $Q^i_j$ is an orthogonal matrix (i.e. $\sum_j^N Q^i_jQ^k_j =\delta^{ik}$), $P^i_j$ is a signed permutation matrix (i.e. a matrix that contains a single nonzero entry in each row and each column, with that nonzero entry being $\pm 1$), $A^i_j$ an arbitrary invertible matrix, and $c^i$ a vector the same size as $\theta^i$.
Note that all these groups possess $T\left(N\right)$ as a normal subgroup corresponding to the parameter shift invariance.
Note further that these groups nest as subgroups, namely $B_N \ltimes T(N) \subset E\left(N\right) \subset \text{Aff}\left(N,\mathbb{R}\right)$.

\subsection{Positive Examples}
We see in all the previous examples that unnaturality typically arises from the fact that the gradient of the loss $\frac{\partial L}{\partial \theta^i}$ is a section of the cotangent bundle $T^*\mathcal{M}$, while the parameter updates are sections of the tangent bundle $T\mathcal{M}$. 
These two bundles are isomorphic as vector bundles, but famously there is no natural isomorphism between them; this unnaturality is the source of the unnaturality in the training algorithms. 
As such, using an explicit isomorphism $T^*\mathcal{M}\to T\mathcal{M}$ in the form of a section of $T\mathcal{M}\otimes T\mathcal{M}$, as in the case of natural gradient descent by the inverse of the Fisher information matrix, restores naturality.
This can be easily done if the output of the network is a vector in an inner product space, as this inner product can be pulled back to obtain a section of $T^*\mathcal{M}\otimes T^*\mathcal{M}$, which can then be inverted provided the network is minimally parameterized and the training data is rich enough for this map to be full rank.
The result of this pull-back is a generalized Gauss-Newton matrix \citep{schraudolph2002fast}, which immediately implies that algorithms utilizing the inverses of Gauss-Newton matrices to precondition gradients are likely natural.

\begin{example}[Generalized Gauss-Newton]
Let a neural network be used to represent a function of the form 
\begin{equation*}
y = f_{NN}\left(x,\theta^i\right),
\end{equation*}
where $x$ are input data, $y$ are output data, and $\theta^i$ are network parameters.
Many matrices fall into the category of a \textbf{generalized Gauss-Newton} matrix, which in components take the form
\begin{equation*}
G_{ij} = \frac{1}{|S|} \sum_{\left(x,y\right)\in S} J^\alpha_i M_{\alpha \beta} J^\beta_j
\end{equation*}
where $J^\alpha_i=\frac{\partial f_{NN}^\alpha}{\partial \theta^i}$ is the Jacobian of the neural network with respect to its parameters, $M_{\alpha \beta}$ is a positive semi-definite matrix that doesn't depend on the network parameters, but may depend on values of $x$ and $y$, and the sum is taken over some set $S$ of input and output data $\left(x,y\right)\in S$.
The matrix $G$ considered in this manner is a positive semi-definite bilinear form on $T\mathcal{M}$, i.e. a section of $T^*\mathcal{M}\otimes T^*\mathcal{M}$.
As such, upon a reparameterization, it transforms as 
\begin{equation*}
\bar{G}_{ij} = \sum_{k,l}^N\frac{\partial \theta^k}{\partial \bar{\theta}^i}G_{kl}\frac{\partial \theta^l}{\partial \bar{\theta}^j}.
\end{equation*}
Assuming that $G_{ij}$ is invertible\footnote{In the case where $G_{ij}$ is not invertible, a pseudo-inverse can be used instead.}, it is easy to see that an evolution of the form
\begin{equation*}
\frac{d\theta^i}{d\xi} = - \sum_j^N\left(G^{-1}\right)^{ij}\frac{\partial L}{\partial \theta^j}
\end{equation*}
transforms to
\begin{equation*}
\frac{d\bar{\theta}^i}{d\xi} = - \sum_j^N\left(\bar{G}^{-1}\right)^{ij}\frac{\partial L}{\partial \bar{\theta}^j},
\end{equation*}
which is precisely the evolution computed in the barred parameters directly.
We therefore see that such an algorithm is natural, as expected.
\end{example}

Returning to the motivating example of natural gradient descent, we can verify that the small learning rate flow are indeed the components of a natural transformation.
\begin{example}[Natural Gradient Descent]
In the small learning rate limit, natural gradient descent becomes
\begin{equation*}
\frac{d\theta^i}{d\xi} = -\sum_j^N g^{ij}\frac{\partial L}{\partial \theta^j},
\end{equation*}
where $g^{ij}$ is the inverse of the Fisher information matrix.
As suggested by the placement of the indices, the components of the Fisher information matrix's inverse represent a map $T^*\mathcal{M}\to T\mathcal{M}$, and transform in a corresponding manner.
A diffeomorphism $\theta^i\to \bar{\theta}^i$ yields
\begin{equation*}
\frac{d\bar{\theta}^i}{d\xi} =\sum_j^N \frac{\partial\bar{\theta}^i}{\partial \theta^j}\frac{d\theta^j}{d\xi}=-\sum_{j,k}^N\frac{\partial\bar{\theta}^i}{\partial \theta^j} g^{jk}\frac{\partial L}{\partial \theta^k}=-\sum_{j,k,l}^N\frac{\partial\bar{\theta}^i}{\partial \theta^j} g^{jk}\frac{\partial\bar{\theta}^l}{\partial \theta^k}\frac{\partial L}{\partial \bar{\theta}^l}=-\sum_{l}^N\bar{g}^{il}\frac{\partial L}{\partial \bar{\theta}^l},
\end{equation*}
which is precisely the same flow that we obtain by transforming to the barred parameters first and then computing the evolution, thanks to the transformation law for the inverse of the Fisher information matrix.
Therefore, natural gradient descent is a natural training algorithm, and as such is appropriately named.
\end{example}
To our knowledge, the realization of natural gradient descent as a discrete approximation of a natural transformation is previously unknown, so the naming is a happy accident.
Taking this naturality together with the rapid convergence of Nesterov's accelerated gradient method suggests a family of training algorithms based on differential equations of the form
\begin{equation}
\label{eqn:NNGD}
\frac{d^2\theta^i}{d\xi^2} = -\frac{r}{\xi}\frac{d\theta^i}{d\xi} -\sum_j^N g^{ij}\frac{\partial L}{\partial \theta^j},
\end{equation}
where $r$ is a suitably chosen constant.
This enjoys the dynamical benefits of Nesterov's method detailed by \cite{su2015differential},  similar to the modification to Adam by \cite{dozat2016incorporating}, while still enjoying the naturality of natural gradient descent.
Any numerical integration scheme applied to this equation can then be thought of as a naturalized version of Nesterov's algorithm.
Further, because any section of $T\mathcal{M}\otimes T\mathcal{M}$ provides the same benefits, the evolution equation \ref{eqn:NNGD} could instead use a generalized Gauss-Newton matrix instead of the Fisher information matrix.
We would then have a training algorithm based on the accelerated generalized Gauss-Newton equation:
\begin{equation}
\label{eqn:AGN}
\frac{d^2\theta^i}{d\xi^2} = -\frac{r}{\xi}\frac{d\theta^i}{d\xi} -\sum_j^N \left(G^{-1}\right)^{ij}\frac{\partial L}{\partial \theta^j},
\end{equation}

\section{Summary and Future Steps}
We have shown that certain machine learning algorithms can be considered as discrete approximations of natural transformations, in the sense that in the limit as learning rate goes to zero, these algorithms converge to a natural transformation between functors mapping $\mathsf{NNet}$ to $\mathsf{Man}$. 
Algorithms with this property essentially eliminate the effects of network parameterization from the training behavior \emph{in the infinitesimal learning rate limit}, allowing the design of network architecture to be primarily focused on the representation power of a network without much worry as to how that set is parameterized. 
This allows efficient training of networks with high degrees of parameter covariance however, as noted by \cite{martens2020new}, once a finite step size is introduced, this equivariance breaks down.
Nevertheless, identifying the underlying mathematical structure leading to this group equivariance, i.e. the naturality of the flow, is the first step in constructing geometric integration schemes explicitly designed to preserve this underlying structure at the discrete level.
Further, if this numerical discretization can be done in a functorial way, perhaps through a Lie group integrator, then the naturality construction we use here can be used to prove an analogous equivariance property for the final finite-learning-rate algorithm, essentially eliminating the dependence of the training on the network parameterization.
We note however that the quality of the parameterization is not entirely irrelevant, because in practice these algorithms are performed in finite precision arithmetic, and poor parameterizations will lead to poorly conditioned matrices that must be inverted or used in a least-squares solve.
This conditioning problem provides a more quantitative measure of the suitability of a network's architecture, as the condition number of a matrix can be computed, and the eigenvectors corresponding to small eigenvalues can be used to inform reparameterizations that improve conditioning.

In this work we defined a category $\mathsf{NNet}$ whose objects were the configuration manifolds of neural networks up to isomorphism, and whose morphisms were auto-diffeomorphisms.
The reparameterization invariance enjoyed by natural gradient descent and generalized Gauss-Newton methods in the small learning rate regime  is then the result of these algorithms being discrete approximations of natural transformations.
We used this framework to generate accelerated training equations \ref{eqn:NNGD} and \ref{eqn:AGN}, the naturality of which yields the parameterization invariance we desire while also capturing the acceleration present in Nesterov's method.

Enlarging the category $\mathsf{NNet}$ by including non-invertible morphisms would allow us to compare the training of non-isomorphic networks with each other.
However it is not immediately obvious how to supplement $\mathsf{NNet}$ with morphisms between distinct objects while retaining a natural training algorithm respecting these new morphisms. 
The most immediately attractive option is to include projection maps $\pi$ such that the training ODEs commute with these projection maps, yielding the following commutative diagram:
\begin{center}
\begin{tikzcd}
\mathcal{M} \arrow{d}{\pi} &\mathcal{S_M} \arrow{r}{\eta_{\mathcal{M}}} \arrow[swap]{d}{\mathcal{S}\left(\pi \right)} & T\mathcal{S_M} \arrow{d}{T\mathcal{S}\left(\pi\right)} \\%
\mathcal{N} &\mathcal{S_N} \arrow{r}{\eta_{\mathcal{M}}}& T\mathcal{S_N}
\end{tikzcd}
\end{center}

A training scheme with this property would guarantee that small networks train as similarly as possible to any larger networks containing them as subnetworks.
These networks and projections maps would then form an inverse system, and the training behavior of inverse limits could be used to study the convergence properties of the dynamics of increasingly large networks.
However, due to the nonlinear nature of neural network parameterization, defining these projection maps is nontrivial, likely requiring the use of more sophisticated structures such as fiber bundles, and as such is beyond the scope of this initial work.

While we have characterized reparameterization invariance, other symmetries may also be desirable, such as the invariance under increasing transformations of the objective function described by \cite{ollivier2017information}.
This can be achieved through the monotone rewriting described therein, and can assist in optimizing some non-smooth functions by transforming them into smooth functions with the same minima.
However, even in this case the optimization domain must still possess a smooth structure, as our framework is restricted to continuous optimization problems.
Discrete optimization must be treated differently, because the categories and functors used in this construction (namely the category $\mathsf{Man}$ and the tangent bundle functor $T$) require a smooth structure.

% Acknowledgements and Disclosure of Funding should go at the end, before appendices and references

\acks{The author is grateful for the guidance and encouragement offered by Michael S. Sacks in the pursuit of this line of research. This work was supported by the F32 HL162423-03 grant provided by the NIH.}

% Manual newpage inserted to improve layout of sample file - not
% needed in general before appendices/bibliography.

\bibliography{ref}

\end{document}